\documentclass[11pt]{article}

\usepackage[final]{acl}

\usepackage{times}
\usepackage{latexsym}
\usepackage{paralist}

\usepackage[T1]{fontenc}

\usepackage[utf8]{inputenc}

\usepackage{paralist}
\usepackage{microtype}

\usepackage{inconsolata}

\usepackage{graphicx}
\usepackage{amsmath}
\usepackage{subcaption}

\newenvironment{fontppl}{\fontfamily{ppl}\selectfont}{\par} 
\usepackage{tcolorbox}
\usepackage{listings}
\tcbuselibrary{breakable,skins}
\definecolor{dark}{rgb}{0.2,0.2,0.2}
\definecolor{light}{rgb}{0.85,0.85,0.85}

\newtcolorbox{promptbox}[2][]{
  floatplacement={#2},
  colframe=dark,colback=light!30!white,
  fonttitle=\small\ttfamily,
  fontupper=\small\ttfamily,
  title=#2,
  boxrule=0.5mm, 
  halign=flush left,
}

\definecolor{dark}{HTML}{064a6c}
\definecolor{light}{HTML}{efede1}

\usepackage{booktabs}
\usepackage{adjustbox}

\title{EpiEvolve: Self-Evolving Agents for Streaming Pandemic Forecasting under Regime Shifts}

\author{
Yiming Lu$^{1}$,
Sihang Zeng$^{2}$,
Zhengxu Tang$^{1}$, 
Max Lau$^{1}$,
Fei Liu$^{1}$,
Wei Jin$^{1}$ \\
$^{1}$Emory University \quad $^{2}$University of Washington \\
\texttt{\{yiming.lu, wei.jin\}@emory.edu} \\
}

\begin{document}
\maketitle
\begin{abstract}
Epidemic LLM forecasters are usually trained and evaluated as static supervised models, whereas operational pandemic forecasting is a streaming process in which labels arrive after predictions and disease regimes shift over time. We study this mismatch in weekly COVID-19 hospitalization trend forecasting across five variant regimes. We introduce EpiEvolve, a self-evolving agent that wraps an LLM forecaster trained on the warm-start period and keeps its weights fixed during streaming. EpiEvolve adapts by storing forecast outcomes in a hierarchical episodic memory, reflecting on delayed labels, retrieving cases relevant to the current regime, and distilling recurring errors into strategic rules. The resulting context lets the forecaster reuse its own past predictions and outcomes in later weeks while following a chronological protocol that prevents future leakage. On the streaming dataset, EpiEvolve reaches $0.629$ average accuracy, compared with $0.561$ for the static backbone and $0.325$ for the external CDC ensemble, and reduces recovery lag after regime shifts from $5$ to $2$ weeks. Ablations show that reflection, strategic memory, and regime-aware retrieval each contribute to the gains.


\end{abstract}

\section{Introduction}




Infectious disease forecasting is central to the response to pandemics, where decisions are made under tight time and resource constraints. Useful forecasts must connect epidemiological trajectories with contextual signals about geography, policy, and viral evolution. Recent epidemic LLM forecasters demonstrate that language models are a promising reasoning interface for this setting, capable of representing both numerical trends and unstructured textual evidence within a unified prediction pipeline \citep{du2025advancing,gong2025epillm,li2025fine}. However, while these systems establish that LLMs can be adapted to epidemic forecasting, they are usually trained and evaluated as static supervised models.

This static evaluation paradigm misses the central property of real-world forecasting that the environment changes after deployment. Public health forecasts are issued iteratively, ground truth arrives only after the predictions are fixed, and the data generating process can shift when new pathogen lineages, immunity profiles, or policy conditions emerge \citep{cramer2022evaluation,reich2022collaborative}. 
COVID-19 made this instability visible as successive variants altered transmission and immune escape across regions, so evidence that was predictive in one period could become less reliable in the next \citep{pham2025large,wang2023alarming}. Consequently, a static model trained on a historical window can fail even when it receives rich inputs. Without mechanisms to adapt to or learn from the errors, these failures may persist. The critical question is whether an LLM forecaster can recover once the epidemic enters a shifted epidemiological regime.



Answering this question requires changes to both evaluation and modeling. On the evaluation side, delayed labels impose a strict chronological protocol. A method must issue forecasts using only information available at that week, and it may update its state after the corresponding labels arrive. This constraint prevents future leakage and makes post-shift performance recovery measurable. On the modeling side, regime shifts change the relevance of historical evidence. Recent errors may be informative after a shift, while older examples can mislead even when they appear superficially relevant. Streaming fine-tuning can update parameters, and retrieval can surface past cases, but neither by itself specifies how forecast outcomes should be organized, reflected on, and converted into reusable strategies. This motivates an agentic structure that remembers past forecasts, interprets outcome feedback, and chooses evidence for the next prediction.


To address these challenges, we propose EpiEvolve, a self-evolving agent for streaming epidemic forecasting. EpiEvolve treats delayed forecast errors after deployment as the primary signal for adaptation. Instead of updating model parameters, it enables a frozen epidemic LLM backbone to adapt through hierarchical episodic memory, retrieval conditioned on epidemic regimes, and lesson distillation from outcomes. This design converts forecast errors into reusable lessons, allowing the agent to recover from distribution shifts without costly gradient updates.

We evaluate EpiEvolve on weekly COVID hospitalization trend forecasting across five COVID variant regimes, using a chronological streaming setup that reflects delayed outcome feedback in real deployment. To our knowledge, EpiEvolve is the first self-evolving LLM agent for streaming epidemic forecasting under regime shifts. Our contributions are:



\begin{compactenum}[\textbullet]
    \item \textbf{A self-evolving epidemic forecasting agent.} We introduce EpiEvolve, a streaming LLM agent that adapts after deployment through hierarchical episodic memory, regime-conditioned retrieval, and outcome-informed lesson distillation while keeping the forecasting backbone frozen.
    \item \textbf{A streaming evaluation with delayed feedback.} We formulate COVID hospitalization trend forecasting as a chronological prediction problem across recurring variant regimes, measured by both average accuracy and performance recovery after distribution shifts.
    \item \textbf{An empirical analysis of adaptation mechanisms.} We compare EpiEvolve with static forecasting, retrieval-based memory, reflection-based memory, external LLM in context forecasting, and streaming fine-tuning. EpiEvolve improves average accuracy from 0.561 to 0.629 over a static frozen backbone and reduces recovery lag from 5 to 2 weeks, while ablations isolate the contribution of memory-based self-evolution.
\end{compactenum}


\section{Related Work}

\noindent\textbf{Epidemic Forecasting, Pandemic LLMs, and Regime Shifts.}
Operational epidemic forecasting combines iterative prediction with delayed scoring, while the biological, behavioral, and policy context shifts between cycles. The U.S. COVID-19 Forecast Hub and collaborative hubs document both the value of ensembles and the instability of individual models \citep{cramer2022evaluation,reich2022collaborative,howerton2023evaluation}. This is a form of concept drift \citep{gama2014survey} now flagged from neural-representation statistics \citep{greco2025unsupervised,ayers2025detecting}, yet in epidemic streams the cues often surface first in text, such as variant announcements and policy updates. Recent epidemic LLM forecasters unify these signals in a single language interface \citep{du2025advancing,gong2025epillm,li2025fine}, and a parallel line uses LLM agents as planning or policymaking assistants under outbreak scenarios \citep{mao2025epiplanagent,shi2026coordinated,aoki2026ai}. Across these settings, however, methods are trained or prompted on a historical window or operate inside a simulator, leaving open how a deployed forecaster should adapt to its own delayed errors across regime shifts.

\noindent\textbf{Self-Evolving LLM Agents and Memory.}
An emerging line of LLM agent research treats accumulated experience as the unit of adaptation, replacing test-time parameter updates \citep{wang2025self,hu2025test,chen2026test,zheng2025spurious} with memory of past outcomes. Recent systems consolidate verbal reflections into reusable rules \citep{wu2025meta}, distill agent trajectories into retrievable principles \citep{wu2025evolver}, and benchmark such memory under continuous test-time streams \citep{wei2025evo}. A parallel direction treats the memory store itself as the object of learning \citep{zhang2025memgen,zhang2025memevolve}, including procedural memory from long-horizon trajectories \citep{wang2025mobile} and shared experience across cooperating agents \citep{weng2026group}. These systems mostly target episodic tasks with immediate ground truth; temporal forecasting under delayed labels \citep{csaba2024label} and recurring regime shifts \citep{wu2024streambench} lie outside their evaluation scope, yet the design pattern of a frozen backbone with evolving memory transfers naturally to it.

\noindent\textbf{LLMs and Foundation Models for Time Series.}
A complementary direction adapts LLM and transformer architectures directly to numerical time-series prediction, either by reprogramming continuous inputs for a frozen language backbone \citep{jin2023time}, tokenizing the series into a completion vocabulary \citep{ansari2024chronos}, or augmenting sequence modeling with spatial-aware reinforcement learning \citep{ni2026streasoner} and epidemiology-aware neural ODE dynamics \citep{wan2025earth}. This line designs the backbone and representation rather than post-deployment adaptation; the resulting forecasters are precisely the kind of frozen backbone that a self-evolving agent can wrap.
\section{Problem Setting}
\label{sec:problem_setting}

\subsection{Streaming Forecasting Task}

We study epidemic forecasting as a streaming classification problem over geographical regions. Let $\mathcal{S}$ denote the set of regions and let $t=1,\ldots,T$ index weekly forecasting rounds. For each region $s\in\mathcal{S}$ and week $t$, the model observes $x_{s,t}$, which contains only information available before the forecast is issued, including recent epidemic time series, surveillance signals, policy text, and regional context. The target is a trend label $y_{s,t}\in\mathcal{Y}$ from a finite ordinal label set.

At week $t$, the forecaster uses a deployment memory $\mathcal{M}_t$ 
containing only information available before week $t$, such as past cases, 
textual reflections, or distilled lessons. For state $s$, a retrieval operator 
$R$ converts this memory and the current input $x_{s,t}$ into a non-parametric 
context
\[
    r_{s,t}=R(\mathcal{M}_t,x_{s,t}).
\]
The forecast is then produced as
\begin{equation}
    \hat{y}_{s,t}=f_{\theta_t}\!\left(x_{s,t}, r_{s,t}\right), \qquad
    \hat{y}_{s,t}\in\mathcal{Y}.
    \label{eq:stream_prediction}
\end{equation}
This notation covers three cases: static baselines, where both $\theta_t$ and 
$r_{s,t}$ are fixed; memory-based agents, where adaptation occurs through 
updates to $\mathcal{M}_t$ and retrieval context; and streaming fine-tuning 
baselines, where adaptation occurs through updates to $\theta_t$.



The defining constraint is the delayed feedback. At week $t$, forecasts for all regions must be generated using only current parameters $\theta_t$ and 
deployment memory $\mathcal{M}_t$. Labels are revealed only after the entire 
weekly batch has been fixed:
\begin{equation}
    \mathcal{B}_t=\left\{(x_{s,t},\hat{y}_{s,t},y_{s,t}) : s\in\mathcal{S}\right\}.
\end{equation}
The method may update its state for the next round:
\begin{equation}
    (\theta_{t+1},\mathcal{M}_{t+1})
    =U\!\left(\theta_t,\mathcal{M}_t,\mathcal{B}_t\right).
    \label{eq:delayed_update}
\end{equation}

This ordering prevents leakage within the week by ensuring that the results of week $t$ cannot inform the predictions of other regions in the same week. EpiEvolve satisfies this constraint with a pre-trained backbone, $\theta_{t+1}=\theta_t$, and adapts through non-parametric updates to memory, retrieval, and lessons. Streaming fine-tuning baselines may instead update $\theta_t$ after the weekly batch is complete.

\begin{figure*}
    \centering
    \includegraphics[width=\textwidth]{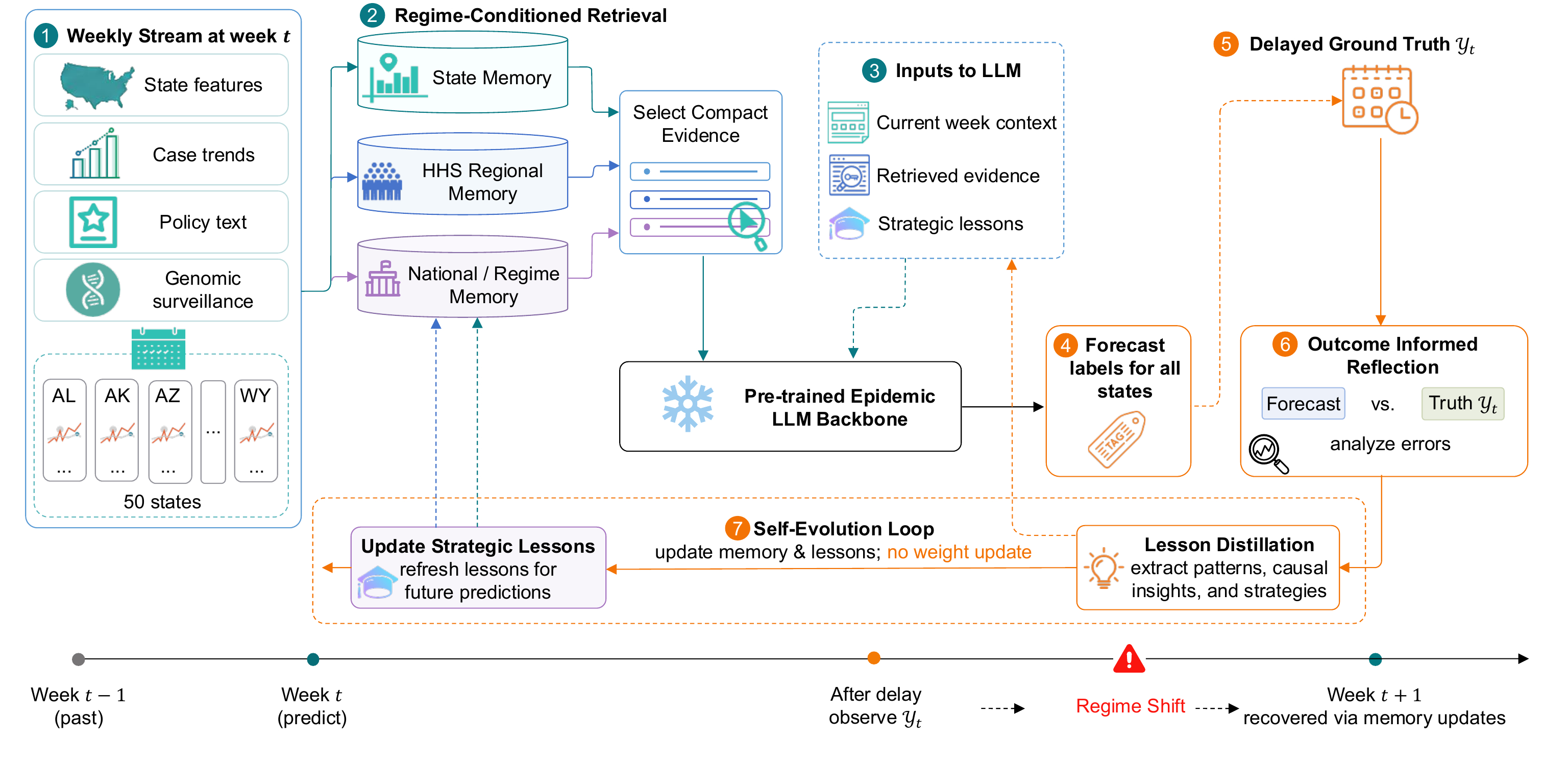}
    \caption{\textbf{Overview of EpiEvolve Pipeline.} Each week $t$, (1) state-level features (epi trends, policy, genomic surveillance) drive (2) regime-conditioned retrieval over episodic memory (state, regional, national), (3) feeding the retrieved cases, distilled lessons, and current context to a pre-trained LLM backbone for (4) hospitalization trend class forecasts. (5) When delayed truth $y_t$ arrives, (6) reflection writes a new episodic entry and triggers lesson distillation. Recovery after regime shifts comes entirely from memory and rule updates.}
    \label{fig:pipeline}
\end{figure*}

\subsection{Regime Evaluation and Data Instantiation}

We instantiate this streaming strategy on weekly COVID-19 hospitalization 
trend forecasting across the 50 U.S. states. Before streaming begins, each 
method receives a warm-start period for offline preparation, such as training 
or initializing the backbone model. Each weekly state-level example combines 
spatial attributes, epidemiological time series, public health policy text, 
vaccination signals, and genomic surveillance information \citep{du2025advancing}. 
The prediction target is a five-class ordinal hospitalization trend label: 
substantial decreasing, moderate decreasing, stable, moderate increasing, or 
substantial increasing. The details of implementation are provided in Section~\ref{sec:exp_setup}.

To assess adaptation under the distribution shift, we partition the streaming period into contiguous regimes, such as dominant variant eras. Let $\mathcal{T}_r$ 
denote the set of weeks in regime $r$. We report regime-level accuracy:
\begin{equation}
    \mathrm{Acc}(r)=
    \frac{1}{|\mathcal{T}_r|\,|\mathcal{S}|}
    \sum_{t\in\mathcal{T}_r}\sum_{s\in\mathcal{S}}
    \mathbf{1}\!\left[\hat{y}_{s,t}=y_{s,t}\right].
    \label{eq:regime_accuracy}
\end{equation}

These regimes are post-hoc partitions used only for reporting and are never supplied to the forecaster. The variant text that appears in $x_{s,t}$ is part of the observable input at forecast time, sourced from public surveillance. We additionally examine recovery around regime transitions, since aggregate accuracy may obscure whether a method adapts quickly after a distributional shift.
\section{Method: EpiEvolve}
\label{sec:method}


We introduce EpiEvolve, a self-evolving agent with external memory for epidemic forecasting. Figure~\ref{fig:pipeline} illustrates the overall framework. We organize the discussion as follows: Section~\ref{sec:method_backbone} introduces the forecasting backbone. Sections~\ref{sec:method_episodic} through~\ref{sec:method_strategic} detail the hierarchical episodic memory, the reflection module responsible for memory updates, and the strategic distillation process. Section~\ref{sec:method_drift} describes the drift detector and the retrieval policy. Finally, Section~\ref{sec:method_loop} integrates these components into a complete streaming pipeline.

\subsection{Forecasting Backbone}
\label{sec:method_backbone}

EpiEvolve instantiates the streaming protocol of Section~\ref{sec:problem_setting} by fixing the backbone parameters and routing all adaptation through the deployment memory $\mathcal{M}_t$. We fine-tune an LLM-based forecaster $f_\theta$ on the warm-start period and keep its parameters fixed throughout streaming evaluation, so $\theta_{t+1} = \theta_t$ at every cycle of Equation~\ref{eq:delayed_update}; the architecture and training procedure are described in Section~\ref{sec:exp_setup}. Withholding gradient updates at deployment is a deliberate design choice. It isolates memory adaptation as the sole driver of the improvement we report, and matches scenarios where backbone weights cannot be touched after release.

Therefore, all adaptation enters $f_\theta$ through the prompt. We extend the prompt template with two structured slots, \texttt{<MEMORY>} and \texttt{<RULES>}, whose contents constitute the retrieved context $r_{s,t}$. When both slots are empty, $f_\theta$ reduces to a static baseline; when populated, it conditions on retrieved evidence without changing its parameters.

\begin{table*}[t]
\centering
\small
\begin{adjustbox}{max width=\textwidth}
\begin{tabular}{llccccccc}
\toprule
Method & Adaptation strategy & Delta / early & BA.1 & BA.2 & BA.5 & BQ.1 & Avg. & Recovery lag \\
\midrule
CDC ensemble & None (external) & 0.393 & 0.458 & 0.469 & 0.139 & 0.165 & 0.325 & N/A \\
PandemicLLM & None & 0.587 & 0.713 & 0.642 & 0.431 & 0.456 & 0.561 & 5 \\
Streaming fine-tune & Gradient update & 0.587 & 0.700 & 0.625 & 0.460 & 0.490 & 0.566 & 5 \\
Claude-ICL & In-context only & 0.563 & 0.624 & 0.561 & 0.479 & 0.447 & 0.557 & 5 \\
Retrieval-only & Episodic retrieval & 0.602 & 0.718 & 0.663 & 0.488 & 0.512 & 0.591 & 4 \\
Reflection + episodic & Reflection memory & 0.611 & 0.728 & 0.672 & 0.523 & 0.541 & 0.604 & 3 \\
\textbf{EpiEvolve} & Reflection + strategic memory & \textbf{0.624} & \textbf{0.751} & \textbf{0.687} & \textbf{0.563} & \textbf{0.582} & \textbf{0.629} & \textbf{2} \\
\bottomrule
\end{tabular}
\end{adjustbox}
\caption{\textbf{Main regime-wise streaming forecasting results.} Regime columns report hospitalization trend classification accuracy for each variant-era slice; Avg. aggregates across the full period; Recovery lag measures how quickly each method recovers after regime transitions. Lower recovery lag is better.}
\label{tab:main_results}
\end{table*}

\subsection{Hierarchical Episodic Memory}
\label{sec:method_episodic}

The deployment memory $\mathcal{M}_t = (\mathcal{E}_t,\, \mathcal{L}_t,\, \rho_t)$ combines a hierarchical episodic store of past forecast outcomes, a strategic memory of distilled rules, and a regime indicator from the drift detector. This subsection describes $\mathcal{E}_t$ and the part of the retrieval operator $R$ that draws from it; Sections~\ref{sec:method_strategic} and~\ref{sec:method_drift} describe $\mathcal{L}_t$ and $\rho_t$.

After a regime shift, errors arrive faster than gradients can absorb them, yet cluster systematically across regions with similar epidemiological context. EpiEvolve records each forecast outcome as an episodic entry
\begin{equation}
    e = (s,\,t,\,\phi_{s,t},\,\hat{y}_{s,t},\,y_{s,t},\,\ell_{s,t},\,\rho_t),
    \label{eq:episodic_entry}
\end{equation}
where $\phi_{s,t}$ embeds recent hospitalization and case trends, vaccination summary, dominant-variant indicator, and policy text. The term $\ell_{s,t}$ is a one-sentence reflection from Section~\ref{sec:method_reflect}, and $\rho_t$ is the regime indicator at write time. Relative to a query at state $s$, $\mathcal{E}_t$ partitions into a state view $\mathcal{E}^{\mathrm{S}}_t(s)$ (entries about $s$), a regional view $\mathcal{E}^{\mathrm{R}}_t(s)$ (other states in the same HHS region), and a national view $\mathcal{E}^{\mathrm{N}}_t(s)$ (other HHS regions).

Retrieval pulls the top-$N$ entries from $\mathcal{E}_t$ under a regime-aware similarity score
\begin{equation}
    \mathrm{score}(e,\,s,\,t) = \cos(\phi_{s,t},\,\phi_{e}) \cdot w(\rho_t,\,\rho_e),
    \label{eq:retrieval_score}
\end{equation}
where $w(\rho_t,\rho_e)=1$ if $\rho_t=\rho_e$ and $\alpha$ otherwise, with $\alpha=0.5$ fixed throughout. The selected entries are rendered into the \texttt{<MEMORY>} slot. The three scopes define how the store is organized, not separate retrieval quotas. Early in a new regime, for example, the current state may have no entries from the same regime, so the ranking can fall back to broader regional or national cases that still match the current features. Section~\ref{sec:exp_memory} ablates this design by restricting retrieval to a single tier.

\subsection{Reflection-Driven Memory Writing}
\label{sec:method_reflect}

Episodic entries are useful only when their content is selective and labeled with the correct correction signal. Once the labels for week $t$ arrive in $\mathcal{B}_t$, a reflection module takes $(x_{s,t},\,\hat{y}_{s,t},\,y_{s,t})$ as input and emits both the one-sentence reflection $\ell_{s,t}$ and a candidate rule fragment. The reflection prompt asks the model to identify which features supported the forecast, which should have shifted it toward the true class, and which regime-conditioned cue the error coincides with. The new entry is appended once to $\mathcal{E}_t$ , and the candidate rule is queued for the strategic distiller. The fixed schema keeps the memory store compact and prevents reflection text from dominating later prompts as the stream grows.

\subsection{Strategic Lesson Distillation}
\label{sec:method_strategic}

Episodic entries are concrete but fragile because a single trajectory may not transfer across regions or regimes. We promote recurrent reflection patterns into a  strategic memory $\mathcal{L}_t$ of predicate-form rules
\begin{equation}
    \lambda = (\mathrm{preconds},\,\mathrm{consequent},\,c,\,n,\,\rho_\lambda),
    \label{eq:rule_schema}
\end{equation}
where the preconditions are a conjunction of feature predicates and the consequent names a target class shift. The support count $n$ is the number of forecast weeks in which the preconditions have matched a region-week observation, and the confidence $c \in [0,1]$ is the empirical fraction of those weeks in which the consequent agreed with the truth, updated on each new match as $c \leftarrow (nc + \mathbf{1}[\text{consequent matched truth}])/(n{+}1)$. The regime tag $\rho_\lambda$ records the regime in which the rule was first distilled. New rules enter $\mathcal{L}_t$ with $n$ initialized to the number of distillation-window entries that satisfied the preconditions and $c$ to the corresponding in-window precision. The distiller runs on a sliding window of recent episodic entries every $K$ weeks and on every drift event from Section~\ref{sec:method_drift}. At each forecast call, rules whose preconditions match $x_{s,t}$ with confidence above a threshold are rendered into the \texttt{<RULES>} slot. Partially matched rules appear as soft hints.

\begin{figure*}[t]
    \centering
    \begin{subfigure}[b]{0.63\textwidth}
        \centering
        \includegraphics[width=\textwidth]{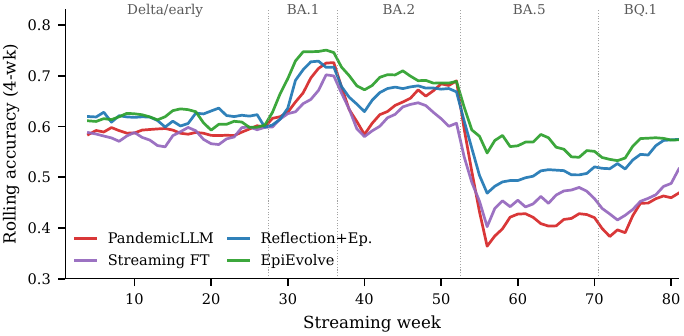}
        \caption{Adaptation curve over the variant-era stream.}
        \label{fig:adaptation}
    \end{subfigure}\hfill
    \begin{subfigure}[b]{0.35\textwidth}
        \centering
        \includegraphics[width=\textwidth]{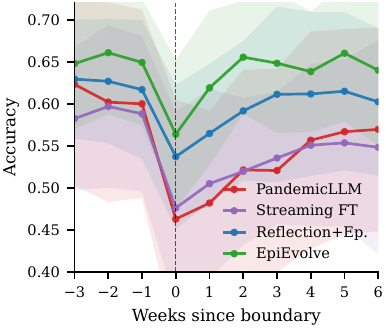}
        \caption{Boundary-centered recovery.}
        \label{fig:recovery}
    \end{subfigure}
    \caption{\textbf{Adaptation behavior across the variant-era stream.} (\subref{fig:adaptation}) Rolling 4-week accuracy across the evaluation period for four representative methods; the static backbone collapses at BA.5 and stays depressed through BQ.1, while EpiEvolve dips less at every boundary and returns to its new-regime steady-state level fastest. (\subref{fig:recovery}) Boundary-centered recovery: weekly accuracy aligned by weeks since each transition and averaged across the four boundaries; shaded bands show across-boundary standard deviation.}
    \label{fig:adaptation_recovery}
\end{figure*}

\subsection{Drift Detector and Retrieval}
\label{sec:method_drift}

Adaptation must respond when the environment changes faster than memory accumulates new evidence. In the post-batch update step, the drift detector evaluates two triggers: (i) the weekly cross-region average ordinal error from $\mathcal{B}_t$ exceeds the warm-start baseline mean by more than $\tau_\sigma$ standard deviations, or (ii) the dominant-variant field in $x_{s,t}$, observable from public surveillance, names a new variant relative to the previous week. Neither trigger reads the evaluation regime labels of Section~\ref{sec:problem_setting}. On a drift event, the regime indicator advances from $\rho_t$ to $\rho_{t+1}$ and strategic distillation runs on the most recent entries of the previous regime. The new $\rho_{t+1}$ biases the retrieval score in Equation~\ref{eq:retrieval_score} for the next forecast cycle and gates $\mathcal{L}_t$ so that obsolete rules are demoted but not deleted.

\subsection{Streaming Agent Loop}
\label{sec:method_loop}

The week-$t$ procedure ties the components together. EpiEvolve enters the week with deployment memory $\mathcal{M}_t$. For every region $s \in \mathcal{S}$, the retrieval operator $R$ assembles $r_{s,t}$ from $\mathcal{E}_t$ and $\mathcal{L}_t$ under $\rho_t$, the backbone produces $\hat{y}_{s,t} = f_\theta(x_{s,t},\, r_{s,t})$, and forecasts for the entire week are emitted before any label is observed, which prevents one region's truth from leaking into another region's prediction in the same week. Once the batched labels $\mathcal{B}_t$ arrive, the reflection module writes new episodic entries, the strategic distiller (if triggered) updates $\mathcal{L}_t$, and the drift detector decides whether to advance the regime indicator from $\rho_t$ to $\rho_{t+1}$, yielding $\mathcal{M}_{t+1}$.
\section{Experiments}
\label{sec:experiments}

We evaluate EpiEvolve as a deployment-time adaptation method for streaming epidemic forecasting. The experiments are organized around three questions: whether memory-based self-evolution improves forecasting under the same chronological information constraint, whether it improves recovery after variant regime transitions, and which mechanisms are responsible for the gains.

\subsection{Experimental Setup}
\label{sec:exp_setup}
\label{sec:exp_main}

\paragraph{Data stream.}
We use a weekly COVID-19 stream that forecasts hospitalization-trend classes for 50 U.S.\ states. The warm-start period covers data through 2021-05-31 and is used for fine-tuning the backbone. Streaming evaluation runs from 2021-06-07 to 2022-12-19, giving 81 weekly rounds and 4{,}050 state-week forecasts. The stream is partitioned into five variant regimes: Late-Alpha to Delta, Omicron BA.1, BA.2, BA.5, and BQ.1.

\paragraph{Metrics.}
The primary metric is hospitalization-trend classification accuracy over the full stream and within each regime. We additionally report mean squared error on the ordinal label space, a boundary recovery curve, and the recovery lag. For each variant boundary at week $t$, let $\bar A_r$ denote the method's mid-regime steady-state accuracy in the new regime $r$, computed as the mean weekly accuracy from week $t+4$ to the regime end (after the boundary dip has subsided). Recovery lag is the smallest $k \geq 1$ for which the weekly accuracy first reaches $\bar A_r$. We report the average across the four variant boundaries and mark a method N/A when the new regime exhibits sustained collapse rather than a transient dip, defined as $\bar A_r$ falling below half of the pre-transition rolling 4-week accuracy at any boundary.

\begin{figure*}[t]
    \centering
    \includegraphics[width=\textwidth]{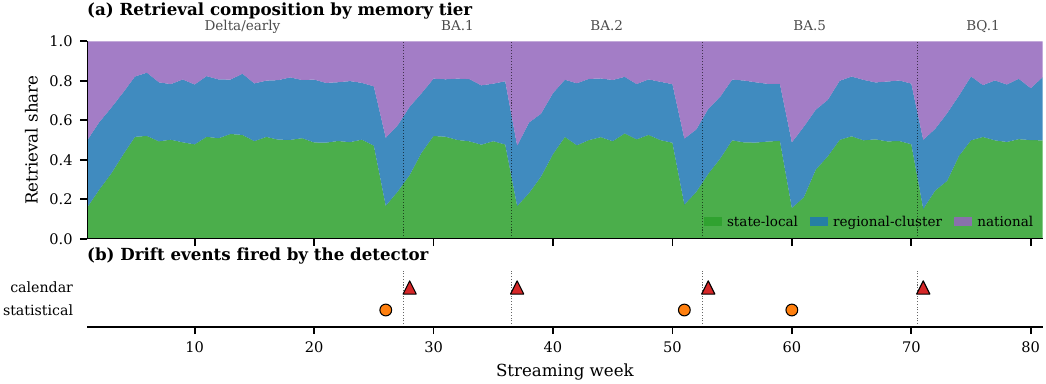}
    \caption{\textbf{EpiEvolve's internal state over the variant regime stream.} Panel (a) shows the memory tier of each top $N$ retrieved entry over time. Within a stable regime, entries from the same state gradually account for more of the retrieved context. At variant transitions, the current regime has few local entries, so retrieval shifts toward regional and national cases with similar features. Panel (b) shows drift events. Triangles denote updates triggered by a change in the dominant variant field of $x_{s,t}$, and circles denote statistical triggers from the rolling error signal.}
    \label{fig:internal_state}
\end{figure*}

\paragraph{Methods.}
All methods obey the delayed feedback protocol and share the same backbone (except Claude-ICL and CDC) fine-tuned on the warm-start period following the PandemicLLM setup of \citet{du2025advancing}. Baselines include PandemicLLM without streaming updates, a streaming fine-tuning baseline with periodic gradient updates, an in-context learning forecaster using Claude, the COVID-19 Forecast Hub ensemble of \citet{cramer2022evaluation} mapped to our five-class hospitalization-trend target, and the retrieval-only and reflection-only variants of EpiEvolve. We evaluate the backbone with Qwen3-14B-Base (Appendix~\ref{sec:method_impl}).

\subsection{Main Results}

Table~\ref{tab:main_results} presents the performance comparison. EpiEvolve outperforms all baselines on both average accuracy and the post-shift regimes where forecasting is hardest. Table~\ref{tab:main_results} reports regime-wise accuracy and recovery lag for each method. PandemicLLM is strong through Delta and BA.1 but collapses when BA.5 emerges in mid-2022, dropping from $0.713$ at BA.1 to $0.431$ at BA.5 and staying below $0.46$ through BQ.1. The external CDC ensemble baseline trails all LLM-based methods, averaging $0.325$ and collapsing below $0.17$ on BA.5 and BQ.1. EpiEvolve attains the highest average accuracy ($0.629$ vs.\ $0.561$) and the lowest recovery lag ($2$ vs.\ $5$ weeks), with its largest margin in the BA.5 and BQ.1 regimes where every other method also struggles. Claude-ICL is comparable to PandemicLLM but below EpiEvolve, indicating EpiEvolve's gain comes from the memory architecture rather than the access to the proprietary LLM.

\subsection{Regime Recovery Analysis}
\label{sec:exp_adaptation}

The time view confirms that EpiEvolve's gains come from post-shift behavior, not from averaging across uneventful weeks. Figure~\ref{fig:adaptation} plots rolling 4-week accuracy across the streaming weeks: PandemicLLM tracks BA.1 and BA.2 well but drops at BA.5 and never recovers within the BA.5 or BQ.1 windows, while EpiEvolve dips less at every boundary and rises back within two weeks. Figure~\ref{fig:recovery} aligns predictions by weeks since each boundary and averages across the transitions; PandemicLLM falls to roughly $0.45$ and barely moves over the next six weeks, while EpiEvolve drops to about $0.58$ and stabilizes at its new-regime steady-state level by week $+2$. Streaming fine-tuning and reflection-only memory sit between these extremes, consistent with their partial gains in Table~\ref{tab:main_results}.

\begin{figure*}[t]
\small
\centering
\setlength{\tabcolsep}{4pt}
\renewcommand{\arraystretch}{1.12}
\begin{tabular}{@{}p{0.97\textwidth}@{}}
\toprule
\textbf{Case study: Florida, week of 2022-06-13 (second week of the BA.5 regime).} \\
\midrule
\textbf{Input} (excerpts; full template in Appendix~\ref{sec:appendix_prompts}). \\
\quad \emph{Variant:} ``BA.5 emerging since 2022-06-06, rapidly displacing BA.2, with higher antibody evasion than prior subvariants.'' \\
\quad \emph{Trend} (last five weeks): \emph{stable, stable, moderate decreasing, stable, stable}. \\
\quad \emph{Dynamic:} Vaccination 67\%; no statewide mask mandate. \\
\quad \texttt{<MEMORY>} (3 of $N{=}8$ entries retrieved by regime-aware score) \\
\quad\quad $\bullet$ FL, 2021-12-20 ($\rho{=}$ BA.1; cross-regime same state): \emph{stable} $\to$ \emph{moderate increasing}. ``BA.1 emergence + vaccination plateau; hospitalizations rose despite two stable weeks.'' \\
\quad\quad $\bullet$ GA, 2022-06-06 ($\rho{=}$ BA.5; regional analogue, HHS-4): \emph{moderate increasing} $\to$ truth confirmed. ``HHS-4 neighbor; BA.5 uptick matched the variant-emergence cue.'' \\
\quad\quad $\bullet$ National, 2022-02-21 ($\rho{=}$ BA.2; cross-regime cross-state): \emph{stable} $\to$ \emph{moderate increasing}. \\
\quad \texttt{<RULES>} (1 of 2 matched rules) \\
\quad\quad $\bullet$ \emph{IF} the variant text mentions ``emerging'' \emph{AND} $\geq 2$ stable weeks in the last five \emph{AND} vaccination $<\!75\%$ \emph{THEN} \emph{moderate increasing}. ($c{=}0.71$, $n{=}14$, regime: variant-shift) \\
\midrule
\textbf{Forecast.} $\hat{y}_{s,t}=$ \emph{moderate increasing} ($p{=}0.62$); truth: \emph{moderate increasing}. Backbone (with empty \texttt{<MEMORY>} and \texttt{<RULES>}): \emph{stable} (incorrect). \\
\midrule
\textbf{Reflection.} ``BA.5 emergence with vaccination near 67\% and two prior stable weeks confirmed the variant-shift uptick.'' \\
\textbf{Rule update.} Matched rule's confidence and support: $c{:}\,0.71\to0.73$, $n{:}\,14\to15$. \\
\bottomrule
\end{tabular}
\caption{\textbf{One forecasting cycle of EpiEvolve walked end to end.} Top block: the model's actual input slots (variant text, recent trend, dynamic features) together with the \texttt{<MEMORY>} and \texttt{<RULES>} populated by hierarchical retrieval and rule matching. Middle block: the model's prediction and a counterfactual from the backbone with empty memory and rules. Bottom block: the agent's writeback for this week, comprising a one-sentence reflection appended to $\mathcal{E}_t$ and accessible to all three memory scopes and the quantitative update to the matched rule's confidence and support.}
\label{fig:case_study}
\end{figure*}

\subsection{Ablation Study}
\label{sec:exp_ablations}
\label{sec:exp_memory}

\begin{table}[t]
\centering
\small
\begin{tabular}{lccc}
\toprule
Configuration & Acc.\ $\uparrow$ & MSE $\downarrow$ & Lag $\downarrow$ \\
\midrule
\textbf{EpiEvolve (full)} & \textbf{0.629} & \textbf{0.65} & \textbf{2} \\
\midrule
\multicolumn{4}{l}{\textit{Component ablations}} \\
\;$-$ strategic memory & 0.604 & 0.71 & 3 \\
\;$-$ reflection (retrieval only) & 0.591 & 0.79 & 4 \\
\;$-$ drift detector & 0.612 & 0.68 & 3 \\
\;$-$ regime-aware retrieval & 0.604 & 0.73 & 3 \\
\midrule
\multicolumn{4}{l}{\textit{Memory scope (single tier)}} \\
\;\;state only & 0.598 & 0.74 & 4 \\
\;\;regional only & 0.602 & 0.70 & 3 \\
\;\;national only & 0.589 & 0.78 & 4 \\
\bottomrule
\end{tabular}
\caption{Component and memory scope ablations. Each configuration removes or replaces one piece of EpiEvolve while keeping the same backbone and delayed feedback protocol.}
\label{tab:ablation}
\end{table}

Table~\ref{tab:ablation} reports two kinds of ablations: removing one component of EpiEvolve at a time, and restricting episodic retrieval to a single tier. The largest accuracy drop comes from removing reflection, which leaves no textual lesson to retrieve and also disables the strategic distiller, collapsing the configuration to a retrieval-only baseline; ordinal MSE rises by $0.14$, indicating that the missing reflection also lets the remaining errors land further from ground truth. Decomposing against the $-0.025$ from removing only strategic memory isolates the reflection text's own contribution to $-0.013$; strategic rules and reflection text therefore split the joint mechanism's gain roughly $2/3$ to $1/3$. Removing strategic memory or restricting retrieval to flat top-$N$ (no regime-aware weighting) each cost between $0.02$ and $0.03$ in accuracy and similar amounts in MSE. The drift detector is different: removing it shifts recovery lag by one week but barely moves MSE, since the agent eventually adapts to the new regime. The matched comparison against streaming fine-tuning under the same delayed feedback constraint shows that prompt and memory adaptation can match or exceed explicit parameter updates while leaving the backbone weights immutable.

Restricting episodic retrieval to a single memory tier costs between $-0.028$ and $-0.041$ in average accuracy, and an MSE penalty of similar magnitude: state-local retrieval misses cross-regional patterns, regional retrieval loses fine-grained local signal, and national retrieval loses geographic specificity. The full hierarchical configuration outperforms every single tier variant on all metrics.

Figure~\ref{fig:internal_state} shows how retrieval and drift detection evolve during the stream. In Panel~(a), state-level entries become more common as a regime matures, since the memory accumulates recent outcomes for the same state. After a transition, those local entries are sparse, and retrieval shifts toward regional and national cases with similar features. Panel~(b) shows that the detector combines two kinds of signals. Four events follow changes in the dominant variant field of $x_{s,t}$, while three additional events are triggered by spikes in rolling ordinal error. These statistical triggers occur at weeks 26, 51, and 60, capturing disruptions that the variant field alone does not mark. This pattern explains why removing the drift detector mainly increases recovery lag in Table~\ref{tab:ablation}. The agent still adapts from later feedback, but it reacts less quickly.

\subsection{Case Study}
\label{sec:case_study}

Figure~\ref{fig:case_study} demonstrates the synergy of our method components (Section~\ref{sec:method}) at a variant boundary. Hierarchical retrieval extracts same-state, regional, and national cross-regime exemplars to capture the ``variant emergence + vaccination plateau $\to$ hospitalizations rise'' pattern, successfully overriding recent local stability. The matched strategic rule reinforces this forecast, while the agent's reflection generates a new insight to inform future retrievals.

\section{Conclusion}

In this paper, we propose EpiEvolve, a self-evolving agent for streaming epidemic forecasting under variant regime shifts that wraps a pre-trained LLM and turns each week's delayed errors into reusable knowledge for the next forecast. The agent uses a hierarchical episodic memory at state, regional, and national scopes, distills recurring patterns into predicate-form rules, and gates retrieval on a regime indicator from a drift detector. Experiments on a multi-regime COVID hospitalization-trend stream show that this self-evolution improves average accuracy and shortens post-shift recovery over baselines, with the largest gains in the most out-of-distribution regimes. Organizing a system's own past errors can therefore substitute for retraining the backbone in this streaming hospitalization-trend setting, suggesting that this style of adaptation may transfer to other forecasting deployments where model weights cannot be touched.

\section*{Limitations}

This work is limited by the scope and granularity of the processed COVID dataset, the unavoidable ambiguity of hard regime boundaries, and the fact that hospitalization trend classification is not a full epidemiological forecasting task. The memory and reflection components may also be sensitive to prompt design, retrieval budget, and the quality of generated strategic lessons. External  LLM baselines may introduce cost and reproducibility concerns.

\section*{Acknowledgments}

\bibliography{custom}

@article{du2025advancing,
  title={Advancing real-time infectious disease forecasting using large language models},
  author={Du, Hongru and Zhao, Yang and Zhao, Jianan and Xu, Shaochong and Lin, Xihong and Chen, Yiran and Gardner, Lauren M and Yang, Hao ‘Frank’},
  journal={Nature Computational Science},
  volume={5},
  number={6},
  pages={467--480},
  year={2025},
  publisher={Nature Publishing Group US New York}
}

@article{cramer2022evaluation,
  title={Evaluation of individual and ensemble probabilistic forecasts of COVID-19 mortality in the United States},
  author={Cramer, Estee Y and Ray, Evan L and Lopez, Velma K and Bracher, Johannes and Brennen, Andrea and Castro Rivadeneira, Alvaro J and Gerding, Aaron and Gneiting, Tilmann and House, Katie H and Huang, Yuxin and others},
  journal={Proceedings of the National Academy of Sciences},
  volume={119},
  number={15},
  pages={e2113561119},
  year={2022},
  publisher={National Academy of Sciences}
}

@misc{reich2022collaborative,
  title={Collaborative hubs: making the most of predictive epidemic modeling},
  author={Reich, Nicholas G and Lessler, Justin and Funk, Sebastian and Viboud, Cecile and Vespignani, Alessandro and Tibshirani, Ryan J and Shea, Katriona and Schienle, Melanie and Runge, Michael C and Rosenfeld, Roni and others},
  journal={American Journal of Public Health},
  volume={112},
  number={6},
  pages={839--842},
  year={2022},
  publisher={American Public Health Association}
}

@article{gong2025epillm,
  title={EpiLLM: unlocking the potential of large language models in epidemic forecasting},
  author={Gong, Chenghua and Sun, Rui and Zheng, Yuhao and Zhang, Juyuan and Gu, Tianjun and Pan, Liming and Lv, Linyuan},
  journal={arXiv preprint arXiv:2505.12738},
  year={2025}
}

@article{li2025fine,
  title={Fine-tuned large language models enhance influenza forecasting},
  author={Li, Chenxiang and Zhang, Qiqiao and Zhang, Yue and Zhao, Bowen and Yang, Jule and Qi, Li and Ding, Jun and Tian, Dechao},
  journal={medRxiv},
  pages={2025--03},
  year={2025},
  publisher={Cold Spring Harbor Laboratory Press}
}

@article{jin2023time,
  title={Time-llm: Time series forecasting by reprogramming large language models},
  author={Jin, Ming and Wang, Shiyu and Ma, Lintao and Chu, Zhixuan and Zhang, James Y and Shi, Xiaoming and Chen, Pin-Yu and Liang, Yuxuan and Li, Yuan-Fang and Pan, Shirui and others},
  journal={arXiv preprint arXiv:2310.01728},
  year={2023}
}

@article{ansari2024chronos,
  title={Chronos: Learning the language of time series},
  author={Ansari, Abdul Fatir and Stella, Lorenzo and Turkmen, Caner and Zhang, Xiyuan and Mercado, Pedro and Shen, Huibin and Shchur, Oleksandr and Rangapuram, Syama Sundar and Arango, Sebastian Pineda and Kapoor, Shubham and others},
  journal={arXiv preprint arXiv:2403.07815},
  year={2024}
}

@article{lewis2020retrieval,
  title={Retrieval-augmented generation for knowledge-intensive nlp tasks},
  author={Lewis, Patrick and Perez, Ethan and Piktus, Aleksandra and Petroni, Fabio and Karpukhin, Vladimir and Goyal, Naman and K{\"u}ttler, Heinrich and Lewis, Mike and Yih, Wen-tau and Rockt{\"a}schel, Tim and others},
  journal={Advances in neural information processing systems},
  volume={33},
  pages={9459--9474},
  year={2020}
}

@article{yao2022react,
  title={React: Synergizing reasoning and acting in language models},
  author={Yao, Shunyu and Zhao, Jeffrey and Yu, Dian and Du, Nan and Shafran, Izhak and Narasimhan, Karthik and Cao, Yuan},
  journal={arXiv preprint arXiv:2210.03629},
  year={2022}
}

@inproceedings{park2023generative,
  title={Generative agents: Interactive simulacra of human behavior},
  author={Park, Joon Sung and O'Brien, Joseph and Cai, Carrie Jun and Morris, Meredith Ringel and Liang, Percy and Bernstein, Michael S},
  booktitle={Proceedings of the 36th annual acm symposium on user interface software and technology},
  pages={1--22},
  year={2023}
}

@article{madaan2023self,
  title={Self-refine: Iterative refinement with self-feedback},
  author={Madaan, Aman and Tandon, Niket and Gupta, Prakhar and Hallinan, Skyler and Gao, Luyu and Wiegreffe, Sarah and Alon, Uri and Dziri, Nouha and Prabhumoye, Shrimai and Yang, Yiming and others},
  journal={Advances in neural information processing systems},
  volume={36},
  pages={46534--46594},
  year={2023}
}

@article{shinn2023reflexion,
  title={Reflexion: Language agents with verbal reinforcement learning},
  author={Shinn, Noah and Cassano, Federico and Gopinath, Ashwin and Narasimhan, Karthik and Yao, Shunyu},
  journal={Advances in neural information processing systems},
  volume={36},
  pages={8634--8652},
  year={2023}
}

@article{gama2014survey,
  title={A survey on concept drift adaptation},
  author={Gama, Jo{\~a}o and {\v{Z}}liobait{\.e}, Indr{\.e} and Bifet, Albert and Pechenizkiy, Mykola and Bouchachia, Abdelhamid},
  journal={ACM computing surveys (CSUR)},
  volume={46},
  number={4},
  pages={1--37},
  year={2014},
  publisher={ACM New York, NY, USA}
}

@article{williams2023epidemic,
  title={Epidemic modeling with generative agents},
  author={Williams, Ross and Hosseinichimeh, Niyousha and Majumdar, Aritra and Ghaffarzadegan, Navid},
  journal={arXiv preprint arXiv:2307.04986},
  year={2023}
}

@article{datta2026agentic,
  title={Agentic Framework for Epidemiological Modeling},
  author={Datta, Rituparna and Guan, Zihan and Espinoza, Baltazar and Su, Yiqi and Pitre, Priya and Venkatramanan, Srini and Ramakrishnan, Naren and Vullikanti, Anil},
  journal={arXiv preprint arXiv:2602.00299},
  year={2026}
}

@article{samaei2026epidemiqs,
  title={EpidemIQs: Prompt-to-paper LLM agents for epidemic modeling and analysis},
  author={Samaei, Mohammad Hosseini and Sahneh, Faryad Darabi and Cohnstaedt, Lee W and Scoglio, Caterina M},
  journal={IEEE Transactions on Artificial Intelligence},
  year={2026},
  publisher={IEEE}
}

@article{mao2025epiplanagent,
  title={EpiPlanAgent: Agentic Automated Epidemic Response Planning},
  author={Mao, Kangkun and Xu, Fang and Ding, Jinru and Jiang, Yidong and Yao, Yujun and Chen, Yirong and Liu, Junming and Wu, Xiaoqin and Wu, Qian and Huang, Xiaoyan and others},
  journal={arXiv preprint arXiv:2512.10313},
  year={2025}
}

@article{aoki2026ai,
  title={AI Agents as Policymakers in Simulated Epidemics},
  author={Aoki, Goshi and Ghaffarzadegan, Navid},
  journal={arXiv preprint arXiv:2601.04245},
  year={2026}
}

@article{shi2026coordinated,
  title={Coordinated Pandemic Control with Large Language Model Agents as Policymaking Assistants},
  author={Shi, Ziyi and Guo, Xusen and Lu, Hongliang and Peng, Mingxing and Wang, Haotian and Zhu, Zheng and Li, Zhenning and Liang, Yuxuan and Zheng, Xinhu and Yang, Hai},
  journal={arXiv preprint arXiv:2601.09264},
  year={2026}
}

@article{ni2026streasoner,
  title={STReasoner: Empowering LLMs for Spatio-Temporal Reasoning in Time Series via Spatial-Aware Reinforcement Learning},
  author={Ni, Juntong and Wang, Shiyu and Jin, Ming and He, Qi and Jin, Wei},
  journal={arXiv preprint arXiv:2601.03248},
  year={2026}
}

@inproceedings{wan2025earth,
  title={EARTH: Epidemiology-Aware Neural ODE with Continuous Disease Transmission Graph},
  author={Wan, Guancheng and Liu, Zewen and Shan, Xiaojun and Lau, Max SY and Prakash, B Aditya and Jin, Wei},
  booktitle={Forty-second International Conference on Machine Learning},
  year={2025}
}

@article{wang2025mobile,
  title={Mobile-agent-e: Self-evolving mobile assistant for complex tasks},
  author={Wang, Zhenhailong and Xu, Haiyang and Wang, Junyang and Zhang, Xi and Yan, Ming and Zhang, Ji and Huang, Fei and Ji, Heng},
  journal={arXiv preprint arXiv:2501.11733},
  year={2025}
}

@article{weng2026group,
  title={Group-Evolving Agents: Open-Ended Self-Improvement via Experience Sharing},
  author={Weng, Zhaotian and Antoniades, Antonis and Nathani, Deepak and Zhang, Zhen and Pu, Xiao and Wang, Xin Eric},
  journal={arXiv preprint arXiv:2602.04837},
  year={2026}
}

@article{howerton2023evaluation,
  title={Evaluation of the US COVID-19 Scenario Modeling Hub for informing pandemic response under uncertainty},
  author={Howerton, Emily and Contamin, Lucie and Mullany, Luke C and Qin, Michelle and Reich, Nicholas G and Bents, Samantha and Borchering, Rebecca K and Jung, Sung-mok and Loo, Sara L and Smith, Claire P and others},
  journal={Nature communications},
  volume={14},
  number={1},
  pages={7260},
  year={2023},
  publisher={Nature Publishing Group UK London}
}

@article{wu2025meta,
  title={Meta-Policy Reflexion: Reusable Reflective Memory and Rule Admissibility for Resource-Efficient LLM Agent},
  author={Wu, Chunlong and Luo, Ye and Qu, Zhibo and Wang, Min},
  journal={arXiv preprint arXiv:2509.03990},
  year={2025}
}

@article{wu2025evolver,
  title={Evolver: Self-evolving llm agents through an experience-driven lifecycle},
  author={Wu, Rong and Wang, Xiaoman and Mei, Jianbiao and Cai, Pinlong and Fu, Daocheng and Yang, Cheng and Wen, Licheng and Yang, Xuemeng and Shen, Yufan and Wang, Yuxin and others},
  journal={arXiv preprint arXiv:2510.16079},
  year={2025}
}

@article{wei2025evo,
  title={Evo-memory: Benchmarking llm agent test-time learning with self-evolving memory},
  author={Wei, Tianxin and Sachdeva, Noveen and Coleman, Benjamin and He, Zhankui and Bei, Yuanchen and Ning, Xuying and Ai, Mengting and Li, Yunzhe and He, Jingrui and Chi, Ed H and others},
  journal={arXiv preprint arXiv:2511.20857},
  year={2025}
}

@article{zhang2025memgen,
  title={Memgen: Weaving generative latent memory for self-evolving agents},
  author={Zhang, Guibin and Fu, Muxin and Yan, Shuicheng},
  journal={arXiv preprint arXiv:2509.24704},
  year={2025}
}

@article{zhang2025memevolve,
  title={Memevolve: Meta-evolution of agent memory systems},
  author={Zhang, Guibin and Ren, Haotian and Zhan, Chong and Zhou, Zhenhong and Wang, Junhao and Zhu, He and Zhou, Wangchunshu and Yan, Shuicheng},
  journal={arXiv preprint arXiv:2512.18746},
  year={2025}
}

@article{pham2025large,
  title={Large-Scale Genomic Analysis of SARS-CoV-2 Omicron BA. 5 Emergence, United States},
  author={Pham, Kien and Chaguza, Chrispin and Lopes, Rafael and Cohen, Ted and Taylor-Salmon, Emma and Wilkinson, Melanie and Katebi, Volha and Grubaugh, Nathan D and Hill, Verity},
  journal={Emerging infectious diseases},
  volume={31},
  number={Suppl 1},
  pages={S45},
  year={2025}
}

@article{wang2023alarming,
  title={Alarming antibody evasion properties of rising SARS-CoV-2 BQ and XBB subvariants},
  author={Wang, Qian and Iketani, Sho and Li, Zhiteng and Liu, Liyuan and Guo, Yicheng and Huang, Yiming and Bowen, Anthony D and Liu, Michael and Wang, Maple and Yu, Jian and others},
  journal={Cell},
  volume={186},
  number={2},
  pages={279--286},
  year={2023},
  publisher={Elsevier}
}

@inproceedings{wang2025self,
  title={Self-updatable large language models by integrating context into model parameters},
  author={Wang, Yu and Liu, Xinshuang and Chen, Xiusi and OBrien, Sean and Wu, Junda and McAuley, Julian},
  booktitle={International Conference on Learning Representations},
  volume={2025},
  pages={16961--16979},
  year={2025}
}

@article{hu2025test,
  title={Test-time learning for large language models},
  author={Hu, Jinwu and Zhang, Zhitian and Chen, Guohao and Wen, Xutao and Shuai, Chao and Luo, Wei and Xiao, Bin and Li, Yuanqing and Tan, Mingkui},
  journal={arXiv preprint arXiv:2505.20633},
  year={2025}
}

@inproceedings{chen2026test,
  title={Test-time adaptation for llm agents via environment interaction},
  author={Chen, Arthur and Liu, Zuxin and Zhang, Jianguo and Prabhakar, Akshara and Liu, Zhiwei and Heinecke, Shelby and Savarese, Silvio and Zhong, Victor and Xiong, Caiming},
  booktitle={The Fourteenth International Conference on Learning Representations},
  year={2026}
}

@article{zheng2025spurious,
  title={Spurious forgetting in continual learning of language models},
  author={Zheng, Junhao and Cai, Xidi and Qiu, Shengjie and Ma, Qianli},
  journal={arXiv preprint arXiv:2501.13453},
  year={2025}
}

@article{greco2025unsupervised,
  title={Unsupervised concept drift detection from deep learning representations in real-time},
  author={Greco, Salvatore and Vacchetti, Bartolomeo and Apiletti, Daniele and Cerquitelli, Tania},
  journal={IEEE Transactions on Knowledge and Data Engineering},
  year={2025},
  publisher={IEEE}
}

@article{ayers2025detecting,
  title={Detecting Concept Drift in Neural Networks Using Chi-squared Goodness of Fit Testing},
  author={Ayers, Jacob Glenn and Ramanan, Buvaneswari A and Khan, Manzoor A},
  journal={arXiv preprint arXiv:2505.04318},
  year={2025}
}

@article{xu2026mem,
  title={A-mem: Agentic memory for llm agents},
  author={Xu, Wujiang and Liang, Zujie and Mei, Kai and Gao, Hang and Tan, Juntao and Zhang, Yongfeng},
  journal={Advances in Neural Information Processing Systems},
  volume={38},
  pages={17577--17604},
  year={2026}
}

@inproceedings{qin2025towards,
  title={Towards adaptive memory-based optimization for enhanced retrieval-augmented generation},
  author={Qin, Qitao and Luo, Yucong and Lu, Yihang and Chu, Zhibo and Liu, Xiaoman and Meng, Xianwei},
  booktitle={Findings of the Association for Computational Linguistics: ACL 2025},
  pages={7991--8004},
  year={2025}
}

@inproceedings{fan2025research,
  title={Research on the online update method for retrieval-augmented generation (rag) model with incremental learning},
  author={Fan, Yuxin and Wang, Yuxiang and Liu, Lipeng and Tang, Xirui and Sun, Na and Yu, Zidong},
  booktitle={2025 5th International Conference on Neural Networks, Information and Communication Engineering (NNICE)},
  pages={1740--1744},
  year={2025},
  organization={IEEE}
}

@article{liu2025pre,
  title={Pre-training Epidemic Time Series Forecasters with Compartmental Prototypes},
  author={Liu, Zewen and Ni, Juntong and Wang, Bohan and Lau, Max SY and Jin, Wei},
  journal={arXiv preprint arXiv:2502.03393},
  year={2025}
}

@article{wu2024streambench,
  title={Streambench: Towards benchmarking continuous improvement of language agents},
  author={Wu, Cheng-Kuang and Tam, Zhi R and Lin, Chieh-Yen and Chen, Yun-Nung and Lee, Hung-yi},
  journal={Advances in Neural Information Processing Systems},
  volume={37},
  pages={107039--107063},
  year={2024}
}

@article{csaba2024label,
  title={Label delay in online continual learning},
  author={Csaba, Botos and Zhang, Wenxuan and M{\"u}ller, Matthias and Lim, Ser-Nam and Elhoseiny, Mohamed and Torr, Philip H and Bibi, Adel},
  journal={Advances in Neural Information Processing Systems},
  volume={37},
  pages={119976--120012},
  year={2024}
}

@article{lin2025se,
  title={Se-agent: Self-evolution trajectory optimization in multi-step reasoning with llm-based agents},
  author={Lin, Jiaye and Guo, Yifu and Han, Yuzhen and Hu, Sen and Ni, Ziyi and Wang, Licheng and Chen, Mingguang and Liu, Hongzhang and Chen, Ronghao and He, Yangfan and others},
  journal={arXiv preprint arXiv:2508.02085},
  year={2025}
}
\nocite{*}

\appendix










\section{EpiEvolve Implementation Details}
\label{sec:method_impl}

We instantiate the backbone with Qwen3-14B-Base, adapted to the streaming task on the warm-start period; backbone training and checkpoint selection are described in Section~\ref{sec:exp_setup}. The agent retrieves $N{=}8$ episodic entries per forecast call from $\mathcal{E}_t$, ranked by the regime-aware score in Equation~\ref{eq:retrieval_score} with cross-regime weight fixed at $\alpha{=}0.5$; the embedding $\phi$ is a 256-dimensional vector from a frozen sentence encoder applied to a structured rendering of $x_{s,t}$. Reflection runs every week; strategic distillation runs every $K{=}4$ weeks and on every drift event; the drift threshold is $\tau_\sigma{=}2$. $\mathcal{E}_t$ is capped and trimmed by a salience score that combines entry recency with the magnitude of the ordinal forecast error, so that informative mistakes are retained longer than uneventful correct predictions. Strategic-rule preconditions are conjunctions of feature comparisons over $x_{s,t}$ fields (numeric thresholds, substring presence, categorical equality). New rules enter $\mathcal{L}_t$ with $n$ and $c$ initialized to the in-window match count and precision over the distillation window, and are pruned when support, confidence, or recency falls below configured thresholds. Episodic entries and rules are persisted as JSON-Lines under \texttt{runs/epievolve/memory/} so that runs are restartable and ablations can replay or perturb the memory state. Sensitivity to $N$, $K$, and $\tau_\sigma$ is reported in the robustness analysis of Section~\ref{sec:exp_robustness}.

\section{Additional Baseline Details}

\paragraph{CDC ensemble.} The COVID-19 Forecast Hub ensemble \citep{cramer2022evaluation} aggregates probabilistic weekly hospitalization forecasts from approximately forty independent teams into a consensus distribution of twenty-three quantile values per (state, week). We use the publicly archived \texttt{COVIDhub-ensemble} forecasts for the streaming window 2021-06-07 to 2022-12-19, restricted to issued dates that precede the corresponding truth week to satisfy the delayed-feedback protocol of Section~\ref{sec:problem_setting}. Each quantile is mapped to a five-class hospitalization-trend label by computing the per-100k weekly rate change against the same 3-week-smoothed baseline used for the ground-truth labels and binning with the fixed thresholds of \citet{du2025advancing}; the final label for each (state, week) is the mode across the twenty-three binned quantiles.

\begin{figure*}[h]
    \centering
    \includegraphics[width=\textwidth]{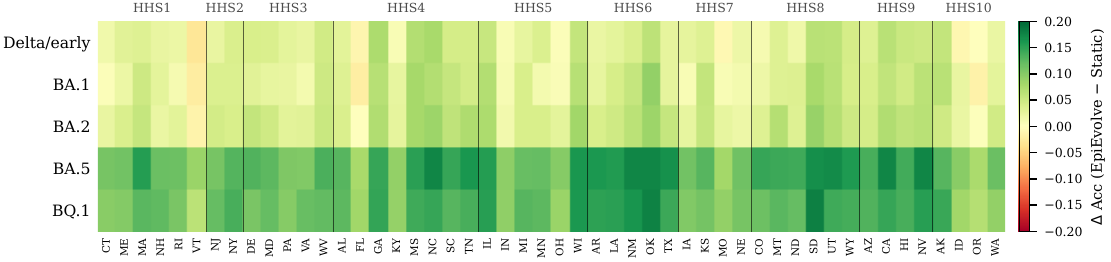}
    \caption{\textbf{Per-state EpiEvolve gain over the backbone.} Each cell is the accuracy of EpiEvolve minus the backbone for one state in one regime; states are grouped by HHS region. Gains concentrate in the BA.5 and BQ.1 rows where the backbone is furthest from its training distribution, and they correlate within HHS regions, since states that share federal coordination structure also share evidence available to EpiEvolve's regional memory tier.}
    \label{fig:per_state_heatmap}
\end{figure*}

\section{Robustness}
\label{sec:exp_robustness}

\begin{table}[h]
\centering
\small
\begin{tabular}{lcc}
\toprule
Configuration & Avg.\ Acc. & Rec.\ lag \\
\midrule
$N = 4$ & 0.612 & 2 \\
$N = 8$ \emph{(default)} & \textbf{0.629} & \textbf{2} \\
$N = 12$ & 0.616 & 2 \\
\midrule
$K = 2$ & 0.622 & 2 \\
$K = 4$ \emph{(default)} & \textbf{0.629} & \textbf{2} \\
$K = 8$ & 0.618 & 3 \\
\midrule
$\tau_\sigma = 1$ & 0.620 & 2 \\
$\tau_\sigma = 2$ \emph{(default)} & \textbf{0.629} & \textbf{2} \\
$\tau_\sigma = 3$ & 0.625 & 3 \\
\bottomrule
\end{tabular}
\caption{Hyperparameter sensitivity. Each block varies one parameter while others remain the default. $N$: total retrieval budget across the three memory tiers. $K$: distillation cadence in weeks. $\tau_\sigma$: drift threshold in standard deviations. Recovery lag is in weeks; lower is better.}
\label{tab:sensitivity}
\end{table}

Table~\ref{tab:sensitivity} varies the three hyperparameters while holding the rest fixed. The results are stable across these settings. Average accuracy ranges from $0.612$ to $0.629$, and recovery lag changes by at most one week. The default values give the best overall tradeoff among the tested settings. Smaller retrieval budgets leave too little evidence in the prompt, less frequent distillation slows rule updates, and stricter drift thresholds delay regime detection. We observe similar stability when shifting each evaluation regime boundary by one or two weeks; regime accuracy and recovery lag remain unchanged.

\section{Qualitative Analysis}
\label{sec:exp_analysis}

Figure~\ref{fig:per_state_heatmap} complements the ablations by breaking the EpiEvolve gain down per (state, regime) cell. The BA.5 and BQ.1 rows are systematically the brightest, with per state improvement averaging around $0.13$. Within HHS correlation reflects the regional memory $\mathcal{E}^{\mathrm{R}}_t$ pooling evidence across states that share a federal coordination structure. A handful of cells are slightly negative, mostly in early regimes where the backbone already does well and the additional memory context offers little upside.

\section{Per-Class Behavior}
\label{sec:appendix_per_class}

\begin{figure}[h]
\centering
\includegraphics[width=\linewidth]{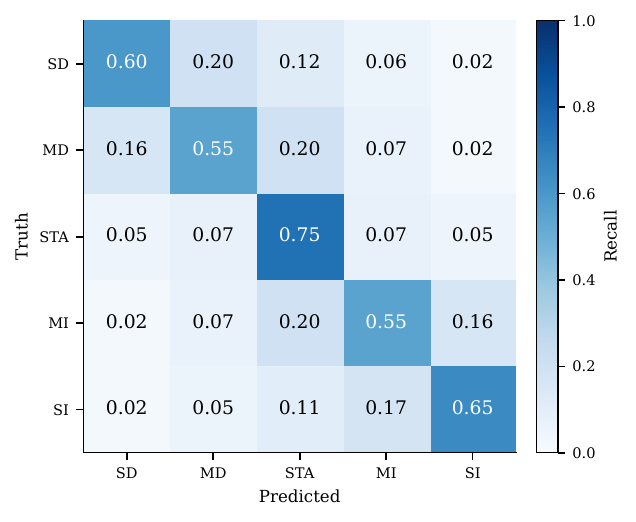}
\caption{\textbf{EpiEvolve confusion matrix (normalized).} Rows are truth classes, columns are predictions.}
\label{fig:confusion_matrix_epievolve}
\end{figure}

To verify that EpiEvolve's gains are broad rather than concentrated in the majority class, Figure~\ref{fig:confusion_matrix_epievolve} reports the row-normalized confusion matrix. The truth label distribution in our streaming window is moderately imbalanced: stable 29\%, moderate increasing 20\%, substantial increasing 18\%, substantial decreasing 17\%, moderate decreasing 16\%. The matrix is diagonal-dominant with off-diagonal mass concentrated on adjacent ordinal classes, and per-class recall stays above 0.55 on every class; the dominant stable class is only slightly easier (0.75) than the harder transition classes (moderate increasing and decreasing at 0.55), indicating that the memory architecture helps the harder classes as much as the easy ones.

\section{Prompt Templates}
\label{sec:appendix_prompts}
EpiEvolve issues three types of LLM calls per week. The backbone $f_\theta$ is fine-tuned on the warm-start period and answers each (state, week) forecast call. Once the delayed labels for week $t$ arrive, the reflection module emits a one-sentence lesson and a candidate rule per (state, week), and the strategic distiller consolidates the most recent reflections into new strategic rules every $K{=}4$ weeks and on each drift event. Bracketed fields are substituted from the agent state at call time.

The lesson is appended to the new episodic entry $e$ of Equation~\ref{eq:episodic_entry}; the candidate rule is queued for the next distillation pass.

\begin{figure*}[t]
\centering
\begin{footnotesize}
\begin{minipage}{\textwidth}

\begin{promptbox}{\fontppl Backbone forecasting prompt: one call per (state{,} week)}

You are an epidemic forecaster. Predict the next week's hospitalization-trend class for one US state.
\begin{lstlisting}
[State]: <state> (FIPS <fips>, HHS region <hhs>)
[Week of]: <target_date>; population <pop>
[Variant]: "<variant_text>"
[Trend] last 5 weeks (oldest first):
  <c_{t-4}>, <c_{t-3}>, <c_{t-2}>, <c_{t-1}>, <c_{t}>
[Dynamic]: vaccinated <vax>%; <policy_text>

<MEMORY> top-N retrieved episodic entries (regime-aware similarity)
  - <state_e>, <date_e> (rho=<regime_e>; <scope>):
    <y_hat_e> -> <y_e>. "<reflection_text>"
  ... (N entries)

<RULES> matched strategic rules; (c, n) = confidence, support
  - IF <precond_conjunction> THEN <class>. (c=<c>, n=<n>, regime=<rho_lambda>)
  ...

OUTPUT
  forecast: <one of: substantial decreasing | moderate decreasing |
            stable | moderate increasing | substantial increasing>
  prob: <float in [0,1]>
  why: <one sentence, <= 25 words>
\end{lstlisting}
\end{promptbox}

\begin{promptbox}{\fontppl Reflection prompt: one call per (state{,} week) after labels arrive}

You are reviewing one weekly forecast post-hoc to extract a reusable lesson.
\begin{lstlisting}
[State]: <state>; [Week of]: <date>; [Regime]: <rho_t>
[Variant]: "<variant_text>"
[Trend used]: <c_{t-4}>, ..., <c_{t}>
[Dynamic]: vaccinated <vax>%; <policy_text>

[Forecast]: <y_hat> (prob <p>) -> [Truth]: <y_true>
[Outcome]: <correct | incorrect, ordinal offset <k>>

Write a one-sentence lesson (<= 30 words) that:
  (1) names features that supported the forecast,
  (2) names features that should have shifted it toward the truth,
  (3) identifies the regime-conditioned cue the error coincides with.

Also propose a candidate rule whose preconditions are a conjunction of 1-4 predicates over {variant text contains "<keyword>", trend pattern, vaccination<%>, policy mention}.

OUTPUT
  lesson: <sentence>
  candidate_rule: IF <conjunction> THEN <class>
\end{lstlisting}
\end{promptbox}
\end{minipage}
\end{footnotesize}
\end{figure*}

\begin{figure*}[t]
\centering
\begin{footnotesize}
\begin{minipage}{\textwidth}
\begin{promptbox}{\fontppl Strategic distillation prompt: every K{=}4 weeks and on each drift event}

You are distilling a sliding window of weekly reflections into strategic rules.
\begin{lstlisting}
[Window]: weeks <t-K+1>..<t> (K=<K>); regime <rho_t>
[Episodic entries] (state, date, forecast -> truth, lesson):
  - <state_1>, <date_1>: <y_hat_1> -> <y_1>. "<lesson_1>"
  - ...
  (<M> total)

[Existing rules in regime <rho_t>] (do not repeat):
  - IF <preconds> THEN <class>. (c=<c>, n=<n>)
  ...

Propose 0 to 3 NEW strategic rules. Each rule must:
  - have 1-4 precondition predicates;
  - be supported by >= 3 entries in the window;
  - differ from every existing rule by >= 1 predicate.

OUTPUT (one per line, or "no_new_rules"):
  - IF <conjunction> THEN <class>.
\end{lstlisting}
\end{promptbox}
\end{minipage}
\end{footnotesize}
\end{figure*}


\end{document}